\documentclass[conference]{IEEEtran}
\IEEEoverridecommandlockouts
\usepackage{comment}
\usepackage{cite}
\usepackage{amsmath,amssymb,amsfonts}
\usepackage{algorithmic}
\usepackage{graphicx}
\usepackage{textcomp}
\usepackage{xcolor}
\def\BibTeX{{\rm B\kern-.05em{\sc i\kern-.025em b}\kern-.08em
    T\kern-.1667em\lower.7ex\hbox{E}\kern-.125emX}}
    
\usepackage{algorithmic}
\usepackage[ruled,vlined,commentsnumbered]{algorithm2e}

\begin{document}

\title{A SELF-TRAINING FRAMEWORK FOR GLAUCOMA GRADING IN OCT B-SCANS\\
\thanks{This work has been partially funded by the following projects: [H2020-ICT-2016-2017, 732613], DPI2016-77869-C2-1-R, AES2017-PI17/00771, AES2017-PI17/00821 and 20901/PI/18. The work of Gabriel Garc\'{i}a has been supported by the State Research Spanish Agency PTA2017-14610-I.}
}

\author{
\IEEEauthorblockN{Gabriel Garc\'ia\IEEEauthorrefmark{1}, Adri\'an Colomer\IEEEauthorrefmark{1}, Rafael Verd\'u-Monedero\IEEEauthorrefmark{2}, Jose Dolz\IEEEauthorrefmark{3} and Valery Naranjo\IEEEauthorrefmark{1}} 
\IEEEauthorblockA{\IEEEauthorrefmark{1}\textit{Instituto de Investigaci\'on e Innovaci\'on en Bioingenier\'ia, I3B}
\textit{Universitat Polit\`ecnica de Val\`encia}, Valencia, Spain.\\
Email: \{jogarpa7, adcogra, vnaranjo\}@i3b.upv.es}
\IEEEauthorblockA{\IEEEauthorrefmark{2}\textit{Departamento de TICs, Universidad Polit\'{e}cnica de Cartagena}, 30202, Cartagena, Spain.}
\IEEEauthorblockA{\IEEEauthorrefmark{3}\textit{Software Engineering, \'Ecole de technologie sup\'erieure}, Montreal, QC H3C 1K3,
Canada.}
}

\maketitle

\begin{abstract}

In this paper, we present a self-training-based framework for glaucoma grading using OCT B-scans under the presence of domain shift. Particularly, the proposed two-step learning methodology resorts to pseudo-labels generated during the first step to augment the training dataset on the target domain, which is then used to train the final target model. This allows transferring knowledge-domain from the unlabeled data. Additionally, we propose a novel glaucoma-specific backbone which introduces residual and attention modules via skip-connections to refine the embedding features of the latent space. By doing this, our model is capable of improving state-of-the-art from a quantitative and interpretability perspective. The reported results demonstrate that the proposed learning strategy can boost the performance of the model on the target dataset without incurring in additional annotation steps, by using only labels from the source examples. Our model consistently outperforms the baseline by 1-3\% across different metrics and bridges the gap with respect to training the model on the labeled target data.

\end{abstract}

\section{Introduction}\label{sec:01_Introduction}

Glaucoma is a degenerative optic neuropathy that causes functional damage and visual field defects by altering several structures in the optic nerve head (ONH) of the retina \cite{weinreb2004}. 
Currently, the diagnostic procedure for detecting glaucoma requires several time-consuming tests, besides a visual examination of medical images, whose interpretation is often subjective for expert ophthalmologists at the grading time \cite{raja2020}. So, many state-of-the-art studies propose image-processing techniques to help experts via machine-learning solutions. 

\begin{figure*}[h]
\begin{center}
\includegraphics[width=17cm]{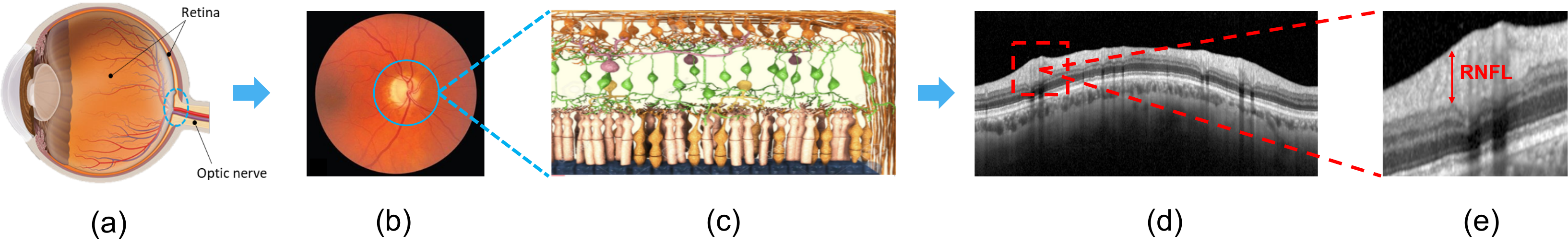}\\
\end{center}
\caption{(a) Eyeball showing the regions of interest. (b) Fundus image highlighting the optic nerve head (ONH). (c) Arrangement of the cell fibre layers of the retina. (d) Typical OCT B-scan evidencing the retinal fibre layers by different grey-intensity levels. (e) Cropping of the RNFL structure.}
\label{OCT_ac}
\end{figure*}

The optical coherence tomography (OCT) is the quintessential imaging modality for glaucomatous damage evaluation since it allows evidencing the deterioration of the cell layers of the optic nerve, which is intimately linked to the glaucoma severity (see Fig. \ref{OCT_ac}). Specifically, the retinal nerve fibre layer (RNFL) has been reported in the clinical literature as the most important structure for glaucoma progression \cite{abd2018}. 

Some of the previous studies combined hand-driven learning algorithms with conventional machine-learning classifiers to discern between healthy and glaucomatous OCT samples \cite{gao2015, kim2017}. The aforementioned studies demand a previous segmentation of the retinal layers of interest to conduct the hand-crafted feature extraction from specific regions of the B-scans, e.g. the RNFL thickness. However, approaches based on prior segmentation knowledge could transfer remaining errors for downstream classification tasks, according to the recent study \cite{thompson2020}. To avoid this shortcoming, deep learning arises as an appealing alternative to derive high-performing computer-aided diagnosis systems in the ophthalmology field. 


Nevertheless, despite the rise of these models in many computer vision and medical problems, their application on OCT images for glaucoma assessment still presents several limitations. First, most of the literature focuses on discerning between healthy and glaucoma classes via OCT B-scans \cite{garcia2020icip, garcia2020ideal}, SD-OCT volumes \cite{garcia2020CMPB, maetschke2019, ran2019} or probability RNFL maps by combining fundus images and OCT samples \cite{thakoor2019, shehryar2020}. Indeed, to the best of our knowledge, only \cite{raja2020} proposes a glaucoma-based scenario beyond the healthy-glaucoma classification by including the suspect label. Furthermore, the limited size of available data sets may hamper the generalization capabilities of learned models, as they could easily lead to overfitting. This is particularly important in the presence of domain shift between training (\textit{labeled}) and testing (\textit{unlabeled}) data distribution. A naive solution would be to collect and labeled data that follows a similar distribution than testing images, which can be then employed to fine-tune the model on the new test data. However, this requires to label large amounts of testing images, involving a time-consuming human process, which is unrealistic in clinical practice. To alleviate the problem of domain shifts, unsupervised domain adaptation has recently appeared as an interesting learning strategy. These methods typically include an adversarial learning framework \cite{yang2020unsupervised, wang2020domain}, which can lead to unstable training and high sensitiveness to hyperparameters.



To fill these gaps in the literature, we propose an alternative learning strategy for glaucoma grading that differs from current literature in several ways. First, unlike existing methods, our approach addresses the problem of glaucoma grading, according to the clinical annotation criteria \cite{flammer1986}. Second, it demonstrates to improve the testing performance of a model trained in the presence of domain shift, approaching the results obtained by full-supervision. Third, the simplicity of our model facilitates the training convergence, contrary to complex adversarial learning-based methods. And last, we propose architectural changes that result in enhanced useful representations from the OCT B-scans, leading to a better performance and more meaningful prediction interpretations compared to conventional architectures.



\section{Methods}\label{sec:02_Methods}

\subsection{Self-training strategy}

\begin{figure}
\begin{center}
\includegraphics[width=9cm]{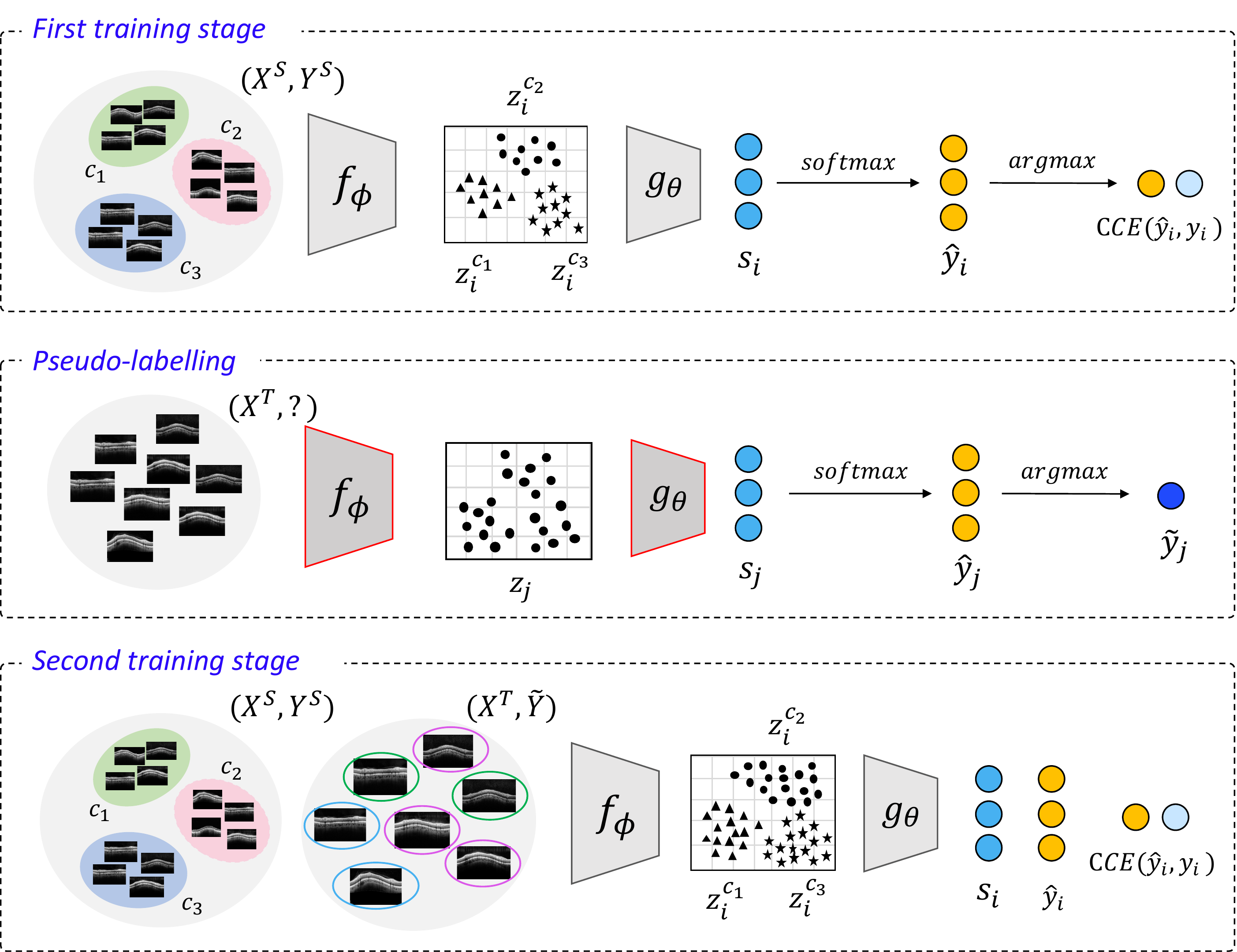}\\
\end{center}
\caption{Illustration of the proposed learning strategy broken down by stages.}
\label{method}
\end{figure}

Self-training, or self-supervised learning, aims at automatically generating a supervisory signal for a given task, which can be then used to enhance the representation learning of features or to label an independent dataset. The former typically involves integrating an unsupervised pretext task \cite{gidaris2018}, relational reasoning \cite{patacchiola2020} or contrastive learning \cite{chen2020}. Nevertheless, in our glaucoma grading scenario, we advocate for the use of a sequential step, where the model is first trained on a labeled source dataset. Then, this model performs an inference on the unlabeled dataset to generate the target pseudo-labels, which are later used to train the model, mimicking full supervision on the test data (see Fig. \ref{method}). This learning strategy has demonstrated a high classification performance on different imaging modalities, such as histopathology \cite{silva2021} or natural images from ImageNet \cite{xie2020}. Formally, we denote $D=\{X, Y\}$ as the independent training set, where $x_i \in X$ refers to the $i$-sample of the database with its corresponding ground truth $y_i \in Y$, being $i=1,2, ..., \mathcal{N}$ the number of training image pairs. Furthermore, we use $T=\{Y, \hat{p}(Y|X)\}$ to represent a given task, where $\hat{p}(Y|X)$ is usually learnt by a neural network. In the current work, we leverage the scenario where the source and target domains are related ($D^S \sim D^T$) since both correspond to OCT samples, but acquired from different hardware settings. Furthermore, tasks across models are the same $(T^S = T^T)$, as we focus on multi-class B-scans classification to discern between healthy, early and advanced glaucoma classes.


Thus, let $x_i \in \mathbb{R}^{M\times N}$ be a raw B-scan of dimensions $M\times N = 248\times 384$, a first model is defined by training a base encoder network $f_\phi(\cdot)$ on the labeled source dataset $D^S$ composed of $\mathcal{N^S}$ samples, where each training instance is denoted by ($x_i$, $y_i$), as observed in Fig. \ref{method}. Then, the embedding representations $z_i=f_\phi(x_i)$ are fed into a classification layer $g_\theta$ to extract the logit scores $s_i$ that are transformed via softmax function to obtain a class-probability $\hat{y}$ (see Algorithm \ref{training}). The coefficients $\phi$ and $\theta$ of the first model are updated during the back-propagation step at every epoch $e=1,2,...,\epsilon$. Once the training is finished on this model, it is used to predict the class of each sample $x_j \in X^T$ from the unlabeled target dataset $D^T$, with $j=1,2, ..., \mathcal{N^T}$, leading to the corresponding pseudo-labels $\tilde{Y}^{T}$ (Algorithm \ref{pseudo_labeling}). Last, the pseudo-labels are used to augment the training dataset, which results in $D^{S,T}=\{X^{S,T}, Y^{S,T}\}$, where $Y^{S,T}={Y}^{S} \cup \tilde{Y}^{T}$. In this way, the model in the last step is trained on the augmented pseudo-labeled dataset, following Algorithm \ref{training}, under the hypothesis that substantial improvements could be reported at the test time because two reasons: i) the increase of labeled training samples and ii) the knowledge distilled from the unlabeled target dataset.

\begin{algorithm}[h]
\caption{Model Training}
\label{training}
\footnotesize
\small
\BlankLine
\KwData{Training set $D^S=\{(x_1,y_1), ..., (x_{\mathcal{N}^S},y_{\mathcal{N}^S})\}$.}
\textbf{Results:} Trained coefficients $\phi, \theta$ \;
\textbf{Algorithm:} \\
$\phi, \theta \leftarrow $ random\;
\For{$ e \leftarrow 1$ \KwTo $\mathbf \epsilon$}{
    \For{$ i \leftarrow 1$ \KwTo $\mathbf \mathcal{N^S}$}{
        $z_i \leftarrow f_{\phi}(x_i)$ \;
        $s_i \leftarrow g_{\theta}(z_i)$ \;
        $\hat{y}_i \leftarrow \frac{\exp{(s_{i})}}{\sum_{c'}{\exp{(-s_{c'})}}} $ \;
    }
    $\mathcal{L}(y,\hat{y}) \leftarrow -\sum_{i}{y_i~log(\hat{y}_i)}$ \;
    Update $\phi$, $\theta$ using $\nabla_{\phi,\theta} \mathcal{L}$ \;
}
\end{algorithm}

\begin{algorithm}[h]
\caption{Pseudo-labeling}
\label{pseudo_labeling}
\footnotesize
\small
\BlankLine
\KwData{Target set $X^T=\{x_1, ..., x_{\mathcal{N}^T}\}$.}
\textbf{Results:} Target pseudo-labels $\tilde{y}$ \;
\textbf{Algorithm:} \\
$\phi, \theta \leftarrow $ frozen\;
\For{$ j \leftarrow 1$ \KwTo $\mathbf \mathcal{N^T}$}{
    $z_j \leftarrow f_{\phi}(x_j)$ \;
    $s_j \leftarrow g_{\theta}(z_j)$ \;
    $\tilde{y}_i \leftarrow \frac{\exp{(s_{j})}}{\sum_{c'}{\exp{(s_{c'})}}} $ \;
}
\end{algorithm}

\subsection{Proposed architecture}

In addition to the presented learning strategy, we propose several architectural changes that improve over existing ones both quantitatively and from an interpretability perspective. In our recent works focused on glaucoma detection from raw OCT samples \cite{garcia2020CMPB, garcia2020icip, garcia2020ideal}, we conducted an in-depth experimental analysis of different deep-learning architectures, both trained from scratch and fine-tuning the most popular models in the literature. Specifically, the VGG family of networks reported the best results in \cite{garcia2020icip}. So, in \cite{garcia2020CMPB}, we employed them as a strong backbone to develop a new feature extractor able to learn useful representations from the slides of a spectral-domain (SD)-OCT volume. Following this findings, in this paper, we adopt our previous work \cite{garcia2020CMPB} as a benchmark to infer the encoder architecture by introducing slight nuances that allow reinforcing the learning of the intrinsic knowledge of OCT samples for glaucoma grading.

\begin{figure*}[t]
\begin{center}
\includegraphics[width=15cm]{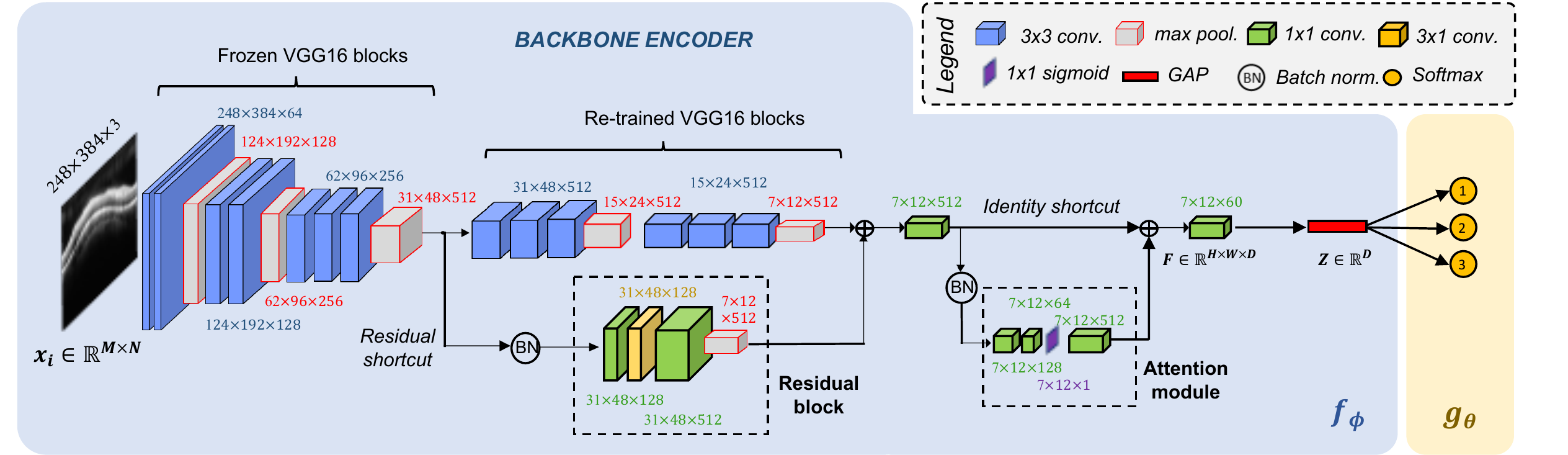}\\
\end{center}
\caption{Backbone architecture used to train the model during the first and second stages. The blue region corresponds to the encoder network structure, whereas the yellow frame denotes the classification layer to discern between healthy, early and advanced glaucomatous cases.}
\label{Backbone_architecture}
\end{figure*}

As observed in Fig. \ref{Backbone_architecture}, we freeze the three first convolutional blocks of the VGG architecture by applying a deep fine-tuning strategy in order to leverage the knowledge acquired by the network when it was pre-trained on the \textit{ImageNet} dataset. Following \cite{garcia2020CMPB}, we include a residual block via convolutional-skip connections and an attention module by means of an identity shortcut to give rise to the $RAGNet\_v2$ architecture. As a novelty, we refine the filters of the residual convolutions to optimize the glaucoma learning process by leveraging the domain-specific knowledge of the OCT samples. In particular, we introduce a tailored kernel size of $3\times 1$ (yellow box) to enforce the network to focus on critical glaucoma-specific regions which underlie contrast changes along the vertical axis of the B-scans. A concatenation aggregation function is used to combine the outputs from the residual block and VGG architecture. Then, a convolution $1\times 1$ is applied to reduce the filters' dimension without affecting the properties of the feature maps. This structure is introduced via skip-connections to refine the embedded space throughout a convolutional autoencoder $1\times 1$ with a sigmoid function aimed to recalibrate the feature learning. Again, a concatenation operation is defined to combine the information from the attention block to the feature map of the main branch. An additional $1\times 1$ convolutional layer is included to provide an embedding volume of features $F$ of dimensions $H\times W\times C = 7\times 12\times 60$, where $C=60$ was empirically calculated as a multiple of the number of classes to encourage a better learning convergence. Finally, a global average-pooling (GAP) layer is applied to compute a spatial squeeze from the feature volume $F$ such that $F\in \mathbb{R}^{H\times W\times C} \rightarrow Z\in \mathbb{R}^C$. In this way, given an input OCT image $x_i \in \mathbb{R}^{M\times N}$, an embedding representation map $z_i \in \mathbb{R}^{C}$ is achieved by the backbone network $f_\phi: X \rightarrow Z$. Regarding the classification stage $g_\theta$, a single output layer with three neurons, corresponding to the number of classes, is implemented through a softmax activation function to determine the probability that $x_i$ belong to the class $c$ (see Fig. \ref{Backbone_architecture}). 


\section{Ablation experiments}\label{sec:03_Ablation_experiments}

\subsection{Data sets}


To evaluate the proposed learning methodology, we resort to two independent databases containing circumpapillary B-scans around the optic nerve head (ONH) of the retina. Note that the OCT samples from both source $D^S$ and target $D^T$ data sets were acquired from different Hospitals using the Heidelberg Spectralis OCT system under distinct settings conditions, e.g. illumination, noisy, contrast, etc. A different senior ophthalmologist (with more than 25 years of clinical experience) annotated each B-scan according to the European Guideline for Glaucoma Diagnosis. We considered $D^T$ as an unlabeled data set during the entire learning process to conduct the proposed methodology. We only used the target labels at the test time to evaluate the models' performance. Information about the data sets distribution per patient and per sample is detailed in Table \ref{dataset_distribution}. Note that the study was conducted according to the guidelines of the Declaration of Helsinki and approved by the Institutional Review Board of each implicated Hospital. Informed consent was obtained from all subjects involved in the study.

\begin{table}[h]
\caption{Patients (pat.) and samples (samp.) per data set grouped by categories, according to the experts' annotation}
\label{dataset_distribution}
\renewcommand{\arraystretch}{1} 
\setlength\tabcolsep{5 pt} 
\centering
\resizebox{8cm}{!}{
\begin{tabular}{cccc|c}
\hline
                    & \begin{tabular}[c]{@{}c@{}}\textbf{Healthy}\\ \textit{(pat./samp.)}\end{tabular} & \begin{tabular}[c]{@{}c@{}}\textbf{Early}\\ \textit{(pat./samp.)}\end{tabular} & \begin{tabular}[c]{@{}c@{}}\textbf{Advanced}\\ \textit{(pat./samp.)}\end{tabular} & \begin{tabular}[c]{@{}c@{}}\textbf{TOTAL}\\ \textit{(pat./samp.)}\end{tabular} \\ \hline
\textbf{Source $D^S$} & 32 / 41       & 28 / 35     & 25 / 31      & 85 / 107     \\
\textbf{Target $D^T$} & 26 / 49       & 24 / 37     & 21 / 26      & 71 / 112     \\ \hline
\textbf{TOTAL}      & 58 / 90       & 52 / 72     & 46 / 57      & 156 / 219    \\ \hline
\end{tabular}}
\end{table}

\textbf{Data partitioning}. At the first stage, we performed a patient-level data partitioning to divide the source data set $D^S$ into five different subsets. A 5-fold cross-validation strategy was addressed to provide robust models and reliable results. In each of the five iterations, $\frac{4}{5}$ of the data were used to train the first model, whereas the remaining samples were employed as a validation subset to prevent overfitting. Otherwise, we randomly selected $\frac{2}{3}$ from the $D^T$ data set to generate the pseudo-labels from which training the model at the second stage. The rest of the target data was used as a test set.

\subsection{Validation of the backbone architecture}
In this stage, we conduct a comparison between the proposed model and the state-of-the-art architectures focusing on OCT-based glaucoma identification. Following the experimental setup carried out in \cite{garcia2020CMPB}, we contrast here the canonical VGG family of networks and the proposed $RAGNet\_v2$ architecture using as a backbone both VGG16 and VGG19 architectures. In Table \ref{val_results}, we show the performance of the aforementioned networks during the training of the model at the first stage in a multi-class scenario for glaucoma grading. To this end, different figures of merit, e.g. sensitivity (SN), specificity (SP), F-score (FS), accuracy (ACC) and area under the ROC curve (AUC), are considered. Note that results correspond to the average and standard deviation from the cross-validation stage in terms of micro-average per class.

\begin{table}[h]
\caption{Micro-average cross-validation results achieved during the training of the first model using the source database $D^S$}
\begin{tabular}{ccccc}
\hline
             & \textbf{VGG16} & \textbf{VGG19} & \textbf{\begin{tabular}[c]{@{}c@{}}RAGNet\_v2\\ (with VGG16)\end{tabular}} & \textbf{\begin{tabular}[c]{@{}c@{}}RAGNet\_v2\\ (with VGG19)\end{tabular}} \\ \hline
\textbf{SN}  & 0.67 $\pm$ 0.06  & 0.75 $\pm$ 0.11  & 0.76 $\pm$ 0.11  & \textbf{0.77 $\pm$ 0.06}   \\
\textbf{SP}  & 0.84 $\pm$ 0.03  & 0.87 $\pm$ 0.05  & 0.88 $\pm$ 0.06  & \textbf{0.89 $\pm$ 0.03}   \\
\textbf{FS}  & 0.67 $\pm$ 0.06  & 0.75 $\pm$ 0.11  & 0.76 $\pm$ 0.11  & \textbf{0.77 $\pm$ 0.06}   \\
\textbf{ACC} & 0.78 $\pm$ 0.04  & 0.83 $\pm$ 0.07  & 0.84 $\pm$ 0.08  & \textbf{0.85 $\pm$ 0.04}   \\
\textbf{AUC} & 0.76 $\pm$ 0.06  & 0.82 $\pm$ 0.08  & 0.82 $\pm$ 0.09  & \textbf{0.83 $\pm$ 0.05}   \\ \hline
\end{tabular}
\label{val_results}
\end{table}

\begin{figure}[h]
\begin{center}
\includegraphics[width=7cm]{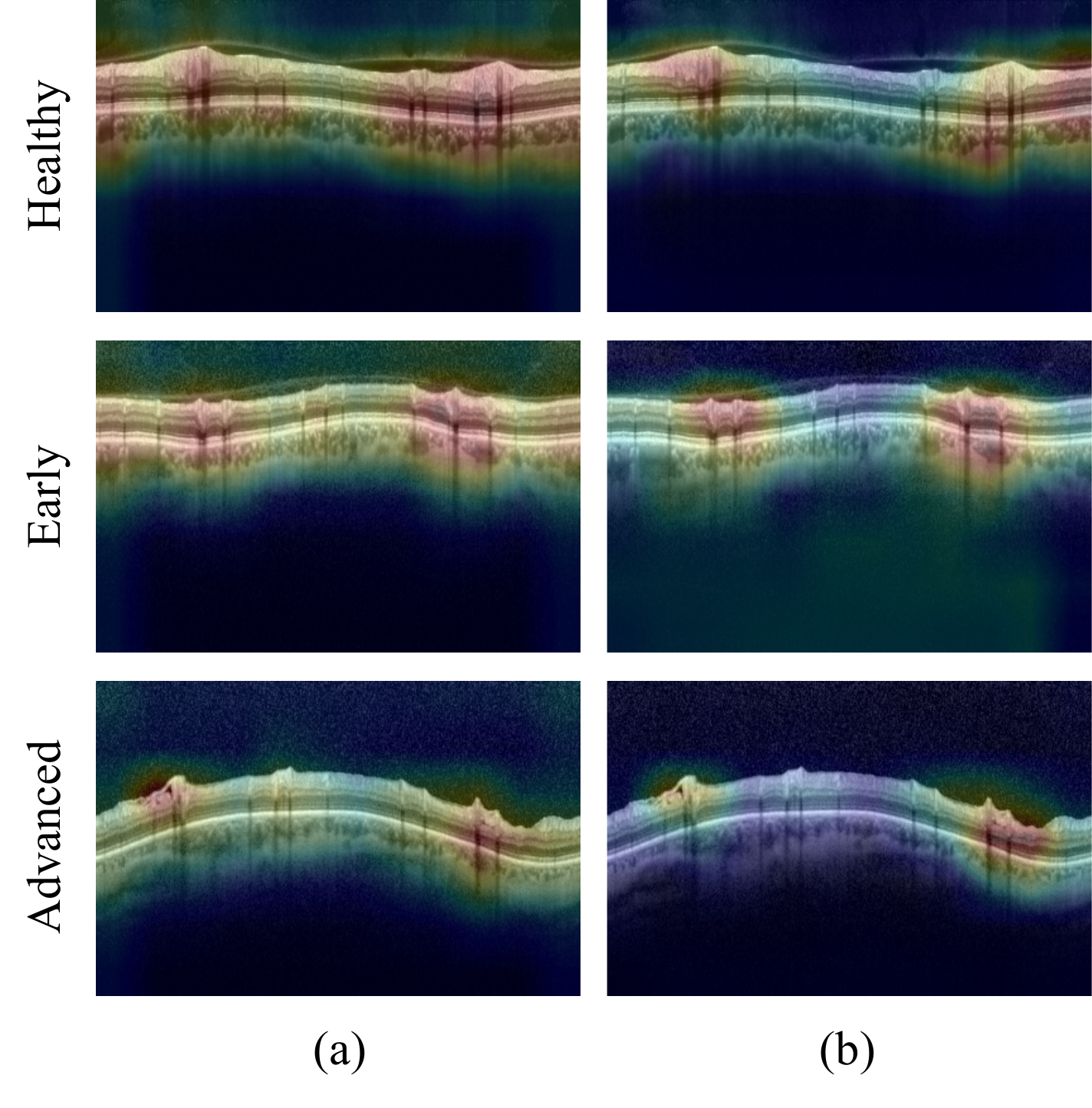}\\
\end{center}
\caption{Class activation maps (CAMs). (a) Heat maps extracted from the VGG19 architecture. (b) Heat maps achieved from the RAGNet\_v2 (with VGG19 as a backbone) at the output from the attention module.}
\label{heatmaps}
\end{figure}

Based on the results from Table \ref{val_results}, we selected the network relative to the $RAGNet\_v2$ (with VGG19) as a baseline to address the pseudo-labeling stage since it outperformed conventional architectures for both VGG16 and VGG19 approaches. In a non-realistic setting, we calculate the performance of the selected backbone at the pseudo-labeling time to determine the usefulness of the proposed approach. To this end, the baseline trained on $(X^S, Y^S)$ was tested on $X^T$, which resulted in accuracy $ACC=0.8376$.  
Besides, the qualitative class activation maps (CAMs) computed in Fig. \ref{heatmaps} further strengthen our confidence in the proposed backbone encoder, since it is evident how the heat maps provided by the attention module (Fig. \ref{heatmaps} (b)) focus on more localized and glaucoma-specific regions than conventional VGG networks (Fig. \ref{heatmaps} (a)). Also, the findings from the CAMs are directly in line with the clinicians' opinion, since the generated heat maps keep an evident relationship between the RNFL thickness and the predicted class, according to the clinical statements \cite{abd2018}. 

\section{Prediction results}\label{sec:04_Prediction_Results}

Once the pseudo-labels $\tilde{y}$ were generated during the first stage, we trained the model making use of the same $RAGNet\_v2$ (with VGG19) architecture. All the experiments were conducted under the same conditions in order to provide a reliable comparison between the different approaches. In particular, all models were trained during 100 epochs using 16 B-scans per batch and the Adadelta optimizer to minimize the categorical cross-entropy (CEE) loss function. At this point, it is important to note that there are no public databases to make a comparison with the literature. In addition, no state-of-the-art studies have been performed to grade the glaucoma severity, so replicating previous glaucoma-based methods would lead to an unreliable and non-objective comparison.


\begin{table*}[t]
\caption{Test results achieved during the prediction of the target set}
\begin{center}
\begin{tabular}{lcccccccccc}
\hline
    & $X^S$ & $X^T$ & $Y^S$ & $Y^T$ & $\tilde{Y}^T$ & \textbf{SN}     & \textbf{SP}     & \textbf{FS}     & \textbf{ACC}    & \textbf{AUC}    \\ \hline
Baseline         & \checkmark &--&\checkmark&--& -- & 0.7059          & 0.8529        & 0.7059          & 0.8039          & 0.7613          \\
Proposed    & \checkmark& \checkmark & \checkmark& -- & \checkmark& \textbf{0.7353} & \textbf{0.8676} & \textbf{0.7353} & \textbf{0.8235} & \textbf{0.7853} \\ \hline
Lower bound      &--&\checkmark&--&--&\checkmark & 0.6765          & 0.8382          & 0.6765          & 0.7843          & 0.7463          \\
Upper bound &  --&\checkmark&--&\checkmark&--& 0.7647          & 0.8824          & 0.7647          & 0.8431          & 0.8050          \\ \hline
\end{tabular}
\label{test_results}
\end{center}
\end{table*}

In Table \ref{test_results}, we report the results achieved by the trained model both in the first stage (\textit{baseline}) and the second stage (\textit{proposed}). Also, as a reference point, we show the performance of the approaches relative to the \textit{upper bound} (model trained with target labels $Y^T$) and the \textit{lower bound} (model just trained with target pseudo-labels $\tilde{Y}^T$). We can observe that the \textit{proposed} learning strategy, which does not require additional target labeled data, consistently outperforms the \textit{baseline} methodology across all the metrics, with improvements of 1-3\%. Note that the \textit{upper bound} scenario is considered to evidence how large the gap in performance is between the fully and semi-supervised approaches. In this case, reported values reveal compelling results, as we observe small differences (2-3\%) between the \textit{upper bound} and the \textit{proposed} strategy. Furthermore, the model just trained on the target dataset making use of the pseudo-labels (\textit{lower bound}) results in a poor-performance with respect to the rest of approaches, as expected, with differences ranging from 3\% to 6\%. This evidences that an increase of the training set via the proposed pseudo-labeling strategy improves the prediction performance for glaucoma grading, as a result of a knowledge transfer between the source $D^S$ and target $D^T$ domains.

\section{Conclusion}\label{sec:05_Conclusion}

The proposed self-training learning strategy has been successfully applied to grade the glaucoma severity from OCT B-scans in the presence of domain shift. Results have demonstrated that including pseudo-labels in the training-loop can enhance the performance over a model trained only on labeled source data, without incurring on extra annotation steps. In addition, the results achieved by the proposed model surpass those reached by the conventional architectures for glaucoma grading, leading to better predictions from both quantitative and interpretability perspective. These findings are evident in the provided heat maps, which highlight more localized glaucoma-specific areas, which are clinically relevant. As a future work, we intend to evaluate our learning strategy across more datasets that might contain larger domain shifts.






\section*{Acknowledgment}

We gratefully acknowledge the support of the Generalitat Valenciana (GVA) for the donation of the DGX A100 used for this work, action co-financed by the European Union through the Programa Operativo del Fondo Europeo de Desarrollo Regional (FEDER) de la Comunitat Valenciana 2014-2020 (IDIFEDER/2020/030).






\bibliographystyle{IEEEtran}
\bibliography{references}


\end{document}